\pdfoutput=1

\documentclass[11pt]{article}

\usepackage{acl}

\usepackage{times}
\usepackage{latexsym}
\usepackage{amssymb}
\usepackage{xcolor}
\usepackage{amsmath}
\usepackage{multirow}
\usepackage{graphicx}
\usepackage{paralist}
\usepackage{booktabs}
\usepackage{array}
\usepackage{multicol}
\usepackage{subcaption}
\usepackage{hyperref}

\usepackage{algorithm,algpseudocode}
\usepackage{newfloat}
\usepackage{listings}

\usepackage[T1]{fontenc}

\usepackage[utf8]{inputenc}

\usepackage{microtype}

\title{ {A}gent-{S}pecific {D}eontic {M}odality {D}etection in {L}egal {L}anguage 
}

\author {
    Abhilasha Sancheti\textsuperscript{$\dagger\ddagger$}, Aparna Garimella\textsuperscript{$\ddagger$}, 
    Balaji Vasan Srinivasan\textsuperscript{$\ddagger$},
    Rachel Rudinger\textsuperscript{$\dagger$} \\
    \textsuperscript{$\dagger$}University of Maryland, College Park\\
    \textsuperscript{$\ddagger$}Adobe Research\\
   \small{\href{mailto:sancheti@umd.edu}{\tt \textcolor{black}{\{sancheti, rudinger\}@umd.edu}}}\\
    \small{\tt \{sancheti, garimell, balsrini\}@adobe.com}
}

\begin{document}
\maketitle
\begin{abstract}
Legal documents are typically long and written in \textit{legalese}, which makes it particularly difficult for laypeople to understand their rights and duties. While natural language understanding technologies can be valuable in supporting such understanding in the legal domain, the limited availability of datasets annotated for deontic modalities in the legal domain, due to the cost of hiring experts and privacy issues, is a bottleneck.
To this end, we introduce, \textsc{LexDeMod}, a corpus of English contracts annotated with deontic modality expressed with respect to a contracting party or agent along with the modal triggers. We benchmark this dataset on two tasks: \begin{inparaenum}[(i)]\item{agent-specific multi-label deontic modality classification, and} \item{agent-specific deontic modality and trigger span detection} \end{inparaenum} using Transformer-based~\citep{vaswani2017attention} language models.
Transfer learning experiments show that the linguistic diversity of modal expressions in \textsc{LexDeMod} generalizes reasonably from lease to employment and rental agreements. A small case study indicates that a model trained on \textsc{LexDeMod} can detect red flags with high recall.
We believe our work offers a new research direction for deontic modality detection in the legal domain\footnote{The code and data are available at \url{https://github.com/abhilashasancheti/LexDeMod}}. 
\end{abstract}

\section{Introduction}
A contract is a legal document executed by two or more parties. 
To sign a contract ({\it e.g.}, lease agreements, terms of services, privacy policies, EULA, etc.), it is important for these parties to precisely understand their obligations, entitlements, prohibitions, and permissions as described in the contract. However, for a layperson, understanding contracts can be difficult due to their length and the complexity of legalese used. Therefore, a layperson often signs agreements without even  reading them~\citep{cole2015rational,obar2020biggest}. Having a system which can provide an ``at a glance'' summary of obligations, entitlements, prohibitions, and permissions to a contracting party (henceforth, ``agent''\footnote{Not related to semantic roles.}), will be of great help not only to the agents but also to legal professionals for contract review. While existing language processing and understanding systems can be used for legal understanding, limited availability of annotated datasets in the legal domain due to the cost of hiring experts and privacy issues is a bottleneck. Furthermore, the highly specialized lexical and syntactic features of legalese make it difficult to directly apply systems trained on data from other linguistic domains (\textit{e.g.}, news) to the legal domain. 
\begin{figure}[t!]
    \centering
    \includegraphics[width=0.40\textwidth,height=7cm]{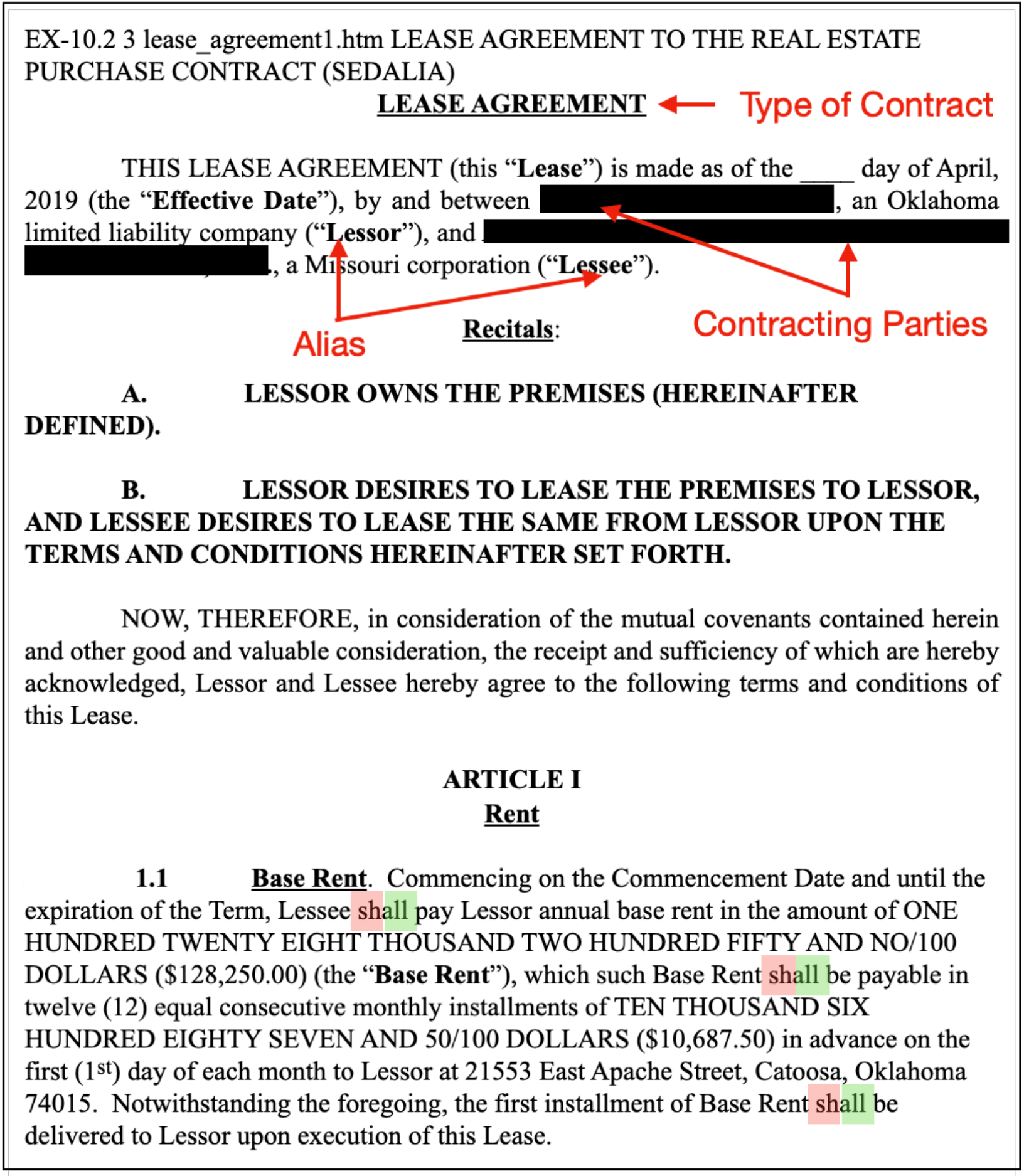}
    \caption{Sample contract\protect\footnotemark~indicating the terminologies used to refer to the elements of a contract. `shall' triggers \colorbox{red!20}{obligation} for Lessee and \colorbox{green!20}{entitlement} for Lessor. Contracting party or agent is referred to via an ``alias'' (such as Lessor or Lessee) throughout the contract.} 
    \label{fig:sample}
\end{figure}
\footnotetext{Party names redacted for anonymity purpose.}

For an ``at a glance'' summary of contracts, we first need to identify the obligations, entitlements, prohibitions, and permissions present in the contract for a given agent. 
Deontic modality is frequently used in contracts to express such obligations, entitlements, permissions, and prohibitions of agents~\citep{ballesteros2016deontic}. For instance, `shall', `shall not', and `may' is used to express `obligation/entitlement', `prohibition', and `permission' respectively in example (1) below.
\begin{enumerate}[(1)]
\item{
\begin{enumerate}[a.]
\item{Tenant \textbf{shall} pay the rent to the Landlord.}
\item{Landlord \textbf{shall not} obtain financing or enter into any agreement affecting the Property.}
\item{Landlord \textbf{may} continue this Lease in effect after Tenant's abandonment and recover Rent as it becomes due.}
\end{enumerate}}
\item{
\begin{enumerate}[a.]
\item{Tenant \textbf{agrees} to pay the rent.} 
\item{Landlord \textbf{is responsible for} maintaining the structural soundness of the house.} 
\end{enumerate}}
\end{enumerate}
However, existing works for identifying such deontic modality types (henceforth ``deontic types'') either use rule-based~\citep{wyner2011rule,peters-wyner-2016-legal,dragoni2016combining,ash2020unsupervised} or data-driven~\citep{neill2017classifying,chalkidis-etal-2018-obligation} approaches, which cannot be directly used for our purpose. This is because rule-based approaches are not robust as they do not (in practice) capture the rich linguistic variety ({\it e.g.}, use of non-modal expressions in (2)) and ambiguity of modal expressions ({\it e.g.}, `shall' in (1a)). Furthermore, annotated datasets used in the data-driven approaches do not consider multiple deontic types for a sentence and their association with the agent ({\it e.g.}, (1a) is an instance of `obligation' for the Tenant and an `entitlement' for the Landlord). Although, \citet{funaki-etal-2020-contract} introduced a corpus with annotations for rights, obligations, and associated agents, it does not cover all the deontic types. Moreover, different corpora consider different deontic types, lacking an accepted standard. 

In this work, we address these issues through the following \textbf{contributions}:
\begin{inparaenum}[(a)]
    \item{we present a linguistically-informed taxonomy for annotating deontic types in the legal domain, and use the taxonomy to build a corpus ({\sc LexDeMod};
    \textsection{\ref{sec:data}}) of English contracts with two types of annotations: \begin{inparaenum}[(i)]\item{all deontic types expressed in a sentence with respect to an agent, and }\item{spans of modal \textit{triggers}, {\it i.e.}, expressions ({\it e.g.}, bold-faced phrases in examples (1) and (2)) that evoke the modal meaning;
    }
    \end{inparaenum}}
  \item{we benchmark the corpus on two tasks: \begin{inparaenum}[(i)]\item{agent-specific multi-label deontic modality classification (\textsection{\ref{sec:eval-ml}}), and }\item{agent-specific deontic modality and trigger span detection (\textsection\ref{sec:eval-bios}) using state-of-the-art Transformer~\citep{vaswani2017attention} models, and }\end{inparaenum}}
 \item{we perform transfer learning experiments (\textsection\ref{sec:transfer}) to investigate the generalizability of diverse modal expressions in \textsc{LexDeMod} and a case study to detect red flags (\textsection\ref{sec:redflag}) in lease agreements.}
\end{inparaenum}
\section{Related Work}
\subsection{NLP in the Legal Domain}
Prior works have investigated a number of tasks in legal NLP domain including legal
judgement prediction~\citep{aletras2016predicting,luo2017learning,zhong2018legal,chen-etal-2019-charge,chalkidis-etal-2019-neural}, legal entity recognition and classification~\citep{cardellino-etal-2017-legal,chalkidis2017extracting,angelidis2018named}, legal question answering~\citep{duan2019cjrc,zhong2020jec}, and legal summarization~\citep{hachey2006extractive,bhattacharya2019comparative,manor-li-2019-plain}. While legal NLP covers a wide range of tasks, limited efforts have been made for contract review despite it being one of the most time-consuming and tedious tasks. \citet{leivaditi2020benchmark} introduced a benchmark for lease contract review for detecting named entities and red flags. \citet{hendrycks2021cuad} introduced a large expert-annotated dataset and \citet{tuggener-etal-2020-ledgar} a large semi-automatically annotated dataset for provision type classification across a variety of contract types. However, these datasets do not contain deontic type annotations which the focus of this work.

\subsection{Rights and Obligation Extraction}
Existing works either propose rule-based methods~\citep{wyner2011rule,peters-wyner-2016-legal} or use a combination of NLP approaches such as syntax and dependency parsing~\citep{dragoni2016combining} for extracting rights and duties from legal documents such as Federal code regulations, European directives or customer protection codes. Another line of works ~\citep{bracewell2014author,neill2017classifying,chalkidis-etal-2018-obligation} use machine learning and deep learning approaches to predict deontic types with the help of small datasets. However, rule-based approaches are not robust due to the rich linguistic variety and ambiguity of modal expressions, and the annotated datasets do not consider multiple deontic types for a sentence and their association with the agents which is important for contract understanding. \citet{matulewska2010deontic} analyzed contracts from different countries and types considering fine-grained deontic modalities covered in them but only considers obligation, permission and prohibition with temporal constraints. \citet{ash2020unsupervised} propose a rule-based unsupervised approach to identify deontic types with respect to an agent and compute statistics for rights and duties for an agent. However, rule-based approaches have limitations as mentioned above. Recently, \citet{funaki-etal-2020-contract} curate an annotated corpus of contracts for recognizing rights and obligations along with the agents using LegalRuleML~\citep{athan2013oasis}. However, the corpus is not publicly available, does not annotate for modal triggers, and does not cover all the deontic types expressed in a contract.

\subsection{Modality Annotation and Detection}
\textit{Modality} refers to the linguistic ability to describe
alternative ways the world could be and is commonly expressed by modal auxiliaries such as \textit{shall}, \textit{will}, \textit{must}, \textit{can}, and \textit{may}. Existing studies have proposed various modality annotation schemas for Portuguese~\citep{hendrickx-etal-2012-modality,avila2015towards} and applied~\citep{quaresma2014automatic} it to build machine learning models to identify the deontic types. However, it does not cover all the deontic types and restrict the identification to three modal auxiliaries. While \citet{athan2013oasis} and \citet{nazarenko-etal-2018-annotation} propose XML-based annotation schema to formally represent legal text in English and highlight the various interpretive issues that arose during the annotation, it does not consider trigger annotation. Although \citet{rubinstein2013toward} and \citet{pyatkin2021possible} consider trigger and modality type (not restricted to modal auxiliaries) annotations at different levels of granularity, fine-grained deontic types as well as association with the agent is not considered. As different studies consider different deontic types lacking an accepted standard, we present a linguistically-informed taxonomy for annotating deontic types and their triggers. 

\section{{\textbf{\textsc{LexDeMod}} Dataset Curation}} \label{sec:data}
We first describe the dataset source (\textsection{\ref{sec:source}}) followed by pre-processing (\textsection{\ref{sec:preprocess}}), annotation protocol (\textsection{\ref{sec:protocol}}), and the quantitative and qualitative analysis (\textsection{\ref{sec:qanalysis}}) of the collected dataset.
\subsection{Dataset Source} \label{sec:source}
We use the contracts available in the LEDGAR corpus~\citep{tuggener-etal-2020-ledgar} which comprises material contracts (Exhibit-10), such as agreements ({\it e.g.}, shareholder/employment/lease/non-disclosure), crawled from Electronic Data Gathering, Analysis, and Retrieval (EDGAR) system. EDGAR is maintained by the U.S. Securities and Exchange Commission (SEC\footnote{https://www.sec.gov/}). The documents filed on SEC are public information and can be redistributed without a further consent.\footnote{https://www.sec.gov/privacy.htm\#dissemination}

\subsection{Contract Pre-processing} \label{sec:preprocess}
The raw contracts in LEDGAR are available in html format. We extract all the paragraphs (henceforth, ``provisions'') from the html (identified by \texttt{<p>} or \texttt{<div>} tags) of a contract, and heuristically filter the provisions defining any terminologies (identified by presence of phrases such as `shall mean', `means', `shall have the meaning', `has the meaning', etc.). As contracts have a hierarchical structure ({\it e.g.}, bullets and sub-bullets), we prepend (see~\ref{sec:combine-bullet}) the higher level context with the lower level ({\it e.g.}, combining sub-bullets with its context in the main bullet). After this, we heuristically extract the type of the contract ({\it e.g.}, lease or employment contract) and the alias ({\it e.g.}, ``Lessee'' in Figure~\ref{fig:sample}) used to refer the contracting parties from the content of the contracts. 

\noindent\textbf{Contract Type Extraction.} We heuristically scan the first $20$\footnote{We found that the structure of contracts is not fixed and table of contents sometimes precedes the actual contract.} provisions to identify the type of the contract using regular expressions (all uppercase characters and presence of `AGREEMENT').

\noindent\textbf{Agent Alias Extraction.} Agent in a contract can be either a person or a company. Therefore, we scan the first $20$ provisions of a contract to find company mentions using lexnlp~\citep{bommarito2021lexnlp}\footnote{lexnlp was better at extracting companies than spaCy.} and named entities with `person' tag  using spaCy~\citep{honnibal2020spaCy} library. We then use regular expression (alias is mentioned in parenthesis (see Figure~\ref{fig:sample}) following the agent mention) to extract the alias used to refer to the found agents in the provisions. For each type of contract, we manually select the most frequently occurring aliases extracted after using the regular expression. 

We collect all the sentences of provisions belonging to a contract wherein alias for an agent is found. We posit that if a sentence does not contain an alias, then deontic type is not expressed for an agent. For instance, `\textit{Any such month-to-month tenancy shall be subject to every other term, covenant and agreement contained herein.}' is a rule and does not specifically mention any deontic type for an agent.

\subsection{Annotation Protocol} \label{sec:protocol}
\noindent\textbf{Annotation task description.} We propose agent-specific deontic modality detection tasks that address the following issues: \begin{inparaenum}[(i)]\item{non-robustness of rule-based extraction of rights and duties as it cannot capture the rich linguistic variety and ambiguity of modal expressions; }\item{lack of standard taxonomy for annotating fine-grained deontic types; }\item{non-association of deontic type with the agent during annotation, and }\item{considering deontic type detection as a single class classification task. }\end{inparaenum} 
Consider, for instance, the following:
\begin{enumerate}[(3)]
\item{\begin{enumerate}[a.]
\item{[\textit{Tenant}] Tenant \textbf{shall}$_{obl}$ pay the rent to the Landlord and \textbf{may}$_{per}$ use the parking space.}
\item{[\textit{Landlord}] Tenant \textbf{shall}$_{ent}$ pay the rent to the Landlord and may use the parking space.}
\end{enumerate}
}
\end{enumerate}
In these examples, the words in bold evoke the modal expression, which we call a \textit{trigger}. For Tenant as the [\textit{Agent}], an obligation (obl) and a permission (per) are expressed in the sentence (3a), and an entitlement (ent) for the Landlord (3b).

Our data collection is performed via crowdsourcing on Amazon Mechanical Turk (AMT).
We ask the workers to provide two types of annotations for each sentence with respect to an agent (referred to via an alias): \begin{inparaenum}[(i)]\item{select all the deontic types expressed, and }\item{select trigger word(s) (as span) for each selected deontic type. }\end{inparaenum}
If a sentence contains more than one agent, we duplicate it to get separate annotations with respect to each agent so that the workers focus their understanding with respect to one agent at a time. 
This task design choice helps in better estimation of the time taken to do each HIT (Human Intelligence Task) as the number of agent mentions in a sentence can vary. This also simplifies the custom annotation interface (see Figure~\ref{fig:interface}) built to get the annotations. Detailed guidelines for annotation are provided in \ref{sec:ann-guide} (Figure~\ref{fig:guidelines}). \\
\noindent\textbf{Taxonomy for deontic type.}
\begin{table}[t!]
\centering
\scriptsize
\begin{tabular}{l l}
\toprule
\textbf{Deontic Type} & \multicolumn{1}{c}{\textbf{Description}}\\
\midrule
Obligation (Obl) & Agent is required to have or do something \\
Entitlement (Ent) & Agent has the right to have or do something \\
Prohibition (Pro) & Agent is forbidden to have or do something\\
Permission (Per) & Agent is allowed to have or do something \\
No Obligation (Nobl) & Agent is not required to have or do something\\
No Entitlement (Nent) & Agent has no right to have or do something\\
\bottomrule
\end{tabular}
\caption{ Taxonomy\protect\footnotemark~for deontic type.}\label{tab:types}
\end{table}
\footnotetext{We also provide an additional option `None' in case none of the deontic types is expressed or it is a rule.} 
We base our taxonomy for deontic type annotation (Table~\ref{tab:types}) on the deontic logic theory of \citet{von1951deontic}. \citeauthor{von1951deontic}'s categorical modals are best suited for legal contracts which talk about rights and duties of contracting parties \citep{ballesteros-etal-2020-severing,matulewska2010deontic} than other categorizations \citep{chung1985tense,palmer2001mood,jespersen2013philosophy} which are not found in contracts (\textit{e.g.}, desiderative, hortative). We also include no-obligation and no-entitlement categories to cover all possible modalities which were found on manual inspection of contracts.\\
\noindent\textbf{Annotation process and requirements.} 
As legalese is syntactically complex and difficult to understand, the annotation task is quite intricate in nature. 
To ensure that the workers properly understand the task, we first conduct a qualification test which explains the task with the help of right and wrong annotation examples along with explanations and contains $10$ questions.
The qualification test is open to workers with $\geq95\%$ approval rate and $\geq1,000$ approved HITs.
Finally, we select $25$ workers who answered all the qualification questions (details in~\ref{sec:qualification}) correctly.

The main annotation task consists of $12$ sentences per HIT, including $2$ quality check questions to ensure workers provide good annotations. We publish $3$ pilot HITs, with revised guidelines in each of them. We also manually check the annotations (selected randomly) to ensure quality and provide feedback to the workers. We observe a learning curve for the task and considerable variation in the time taken per HIT ($7.5\pm1.5$ mins).
After the pilots, the annotations were majorly performed by $3$ workers.
We publish a batch of $50$ HITs with $3$ annotations for each HIT from the $3$ workers.
As we found good inter-annotator agreement between the $3$ workers (see \textsection{\ref{sec:qanalysis}}), we collect only one annotation per HIT for the remaining HITs to get more sentences annotated within reasonable time. 
\subsection{Annotated Dataset Statistics and Analysis} \label{sec:qanalysis}
Each contract contains $202.6(\pm162.4)$ provisions on average (standard deviation in parentheses), with $2.2(\pm1.7)$ sentences per provision; each contract consists of $306.4(\pm235.8)$ sentences on an average.
Among these, $75.8(\pm14.4)$\% of sentences per contract have atleast one agent mentioned in them, with an average length of $65(\pm47)$. 
\begin{table}[t!]
\centering
\resizebox{0.99\columnwidth}{!}{
\begin{tabular}{l c c |c c c c c c c}
\toprule
\textbf{Split} & \textbf{\#Sent.} &\textbf{\#Spans}& \textbf{Obl} & \textbf{Ent} & \textbf{Pro} &\textbf{Per} & \textbf{Nobl} & \textbf{Nent} & \textbf{None}\\
\midrule
Train & $4282$&$5279$&$1841$& $1231$ &$343$ &$289$& $265$ & $239$ & $1071$\\
Dev & $330$ &$421$ & $176$ & $86$ & $20$ & $18$ &$21$ & $22$&$78$\\
Test & $1777$ & $1952$ & $575$&$418$ & $64$ &$167$ & $101$ & $88$ & $539$\\
\bottomrule
\end{tabular}
}
\caption{Dataset Statistics.}\label{tab:data-stats}
\end{table}
\begin{figure}[t!]
    \centering
     \includegraphics[width=\columnwidth,height=4cm]{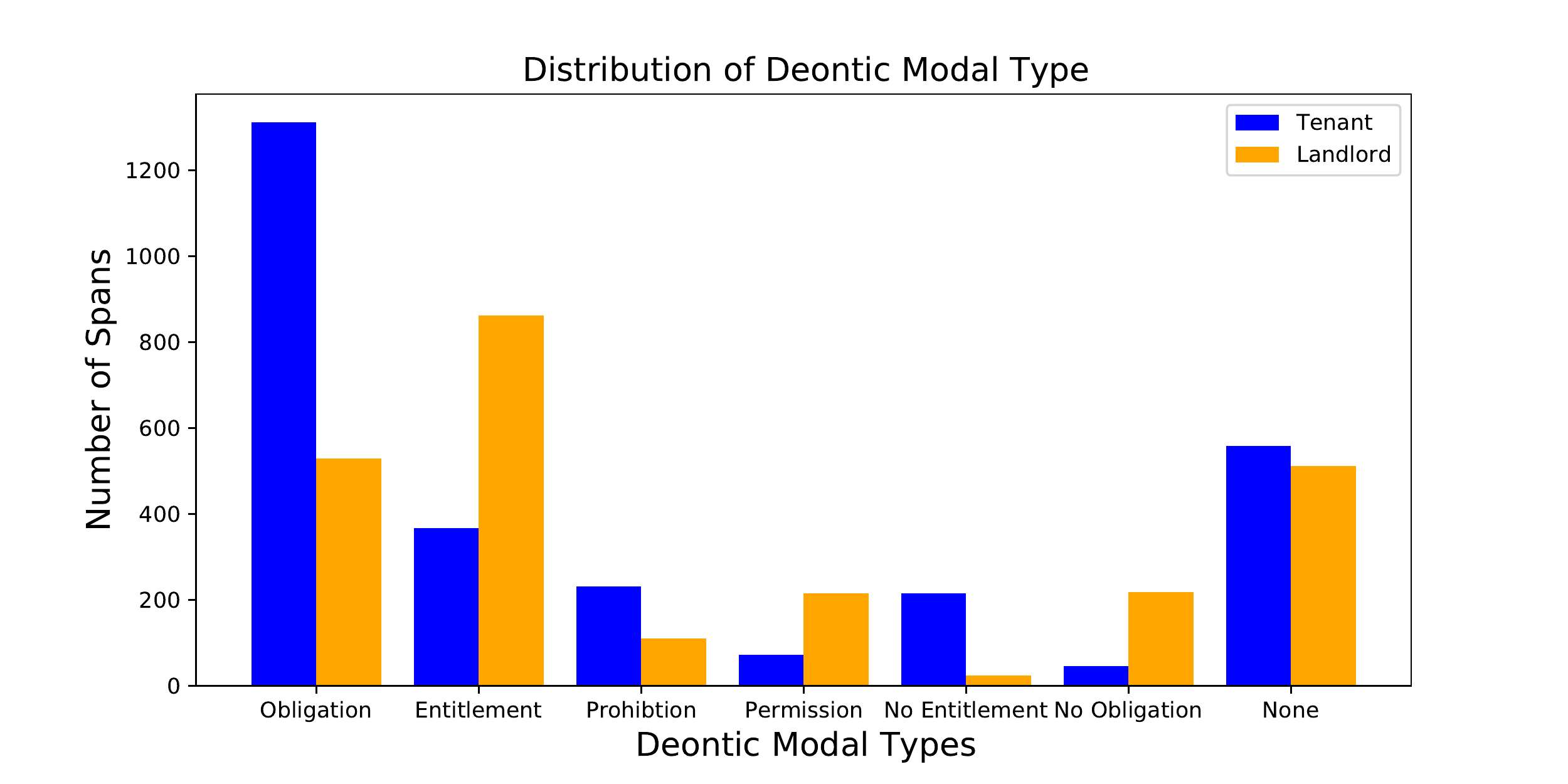}
    \caption{Distribution of deontic type with respect to Tenant and Landlord for lease agreements.}
    \label{fig:distribution}
\end{figure}

We collect a total of $8,230$ trigger span annotations for $7,092$ sentences from $23$ lease contracts after considering HITs for which both the quality questions are correctly answered.
For duplicate sentences, we retain those annotations that are inline with one of the authors (and discard $14.1\%$ of duplicated ones; a few examples are provided in in~\ref{sec:diagreements}). 
The test set comprises of sentences from $5$ contracts including those for which we have $3$ annotations per sentence, and rest of the sentences are divided into train and development sets such that the sentences from the same contract belong to the same set. 
We combine the $3$ annotations for a subset of sentences in the test set using majority voting\footnote{We discard the sentence in case no majority is found.} for deontic type and by taking a union\footnote{Union was performed in $15.54\%$ of sentences and we manually corrected $9.4\%$ of these sentences where union lead to incorrect triggers.} of annotated trigger spans for the majority deontic types. The average inter-annotator agreement for each deontic type computed with Krippendorff's $\alpha$~\citep{krippendorff2018content} is substantial ($\alpha=0.65$) given the complexity of the task (see Table~\ref{tab:agreement} for type-wise agreement). For trigger span annotation, the token-level inter-annotator agreement for the majority deontic types for a sentence is also substantial ($\alpha=0.71$). The fine-grained dataset statistics after filtering and resolving disagreements are presented in Table~\ref{tab:data-stats}. 
\begin{table}[t!]
\centering
\scriptsize
\resizebox{0.99\columnwidth}{!}{
\begin{tabular}{l |p{5cm}}
\toprule
\textbf{Type} & \multicolumn{1}{c}{\textbf{Top 10 triggers}}\\
\midrule
Obl &  shall, will, agrees, agree, acknowledges, acknowledge, represents and warrants, shall be responsible for, undertakes, will be responsible for  \\
Ent & shall, will, agrees, shall have the right to, shall be entitled to, represents and warrants, acknowledges, waives no rights, shall not, retains all other rights, will be entitled to\\
Pro & shall not, will not, may not,  nor shall, not to be, neither lessor nor lessee may, in no event shall, nor will, will not allow, nor may\\
Per & may, is permitted, will allow, has the right, shall, or at landlord's option, shall be permitted to, shall be allowed \\
Nobl & shall not be liable, shall not be obligated to, shall not be required to, shall, shall have no obligation to, in no event shall landlord be obligated to, waives, shall not, shall have no liability\\
Nent & shall, shall have no right to, waives no rights, shall not, shall have no obligation to, waives, shall not be required, shall not be obligated, waive the right, shall not have the right to\\
\bottomrule
\end{tabular}
}
\caption{Top $10$ triggers for each deontic type in decreasing order of frequency.}\label{tab:topktrigger}
\end{table}
\begin{figure}[t!]
    \centering
    \includegraphics[width=0.5\textwidth,height=3cm,trim={1cm 2cm 0cm 2cm},clip]{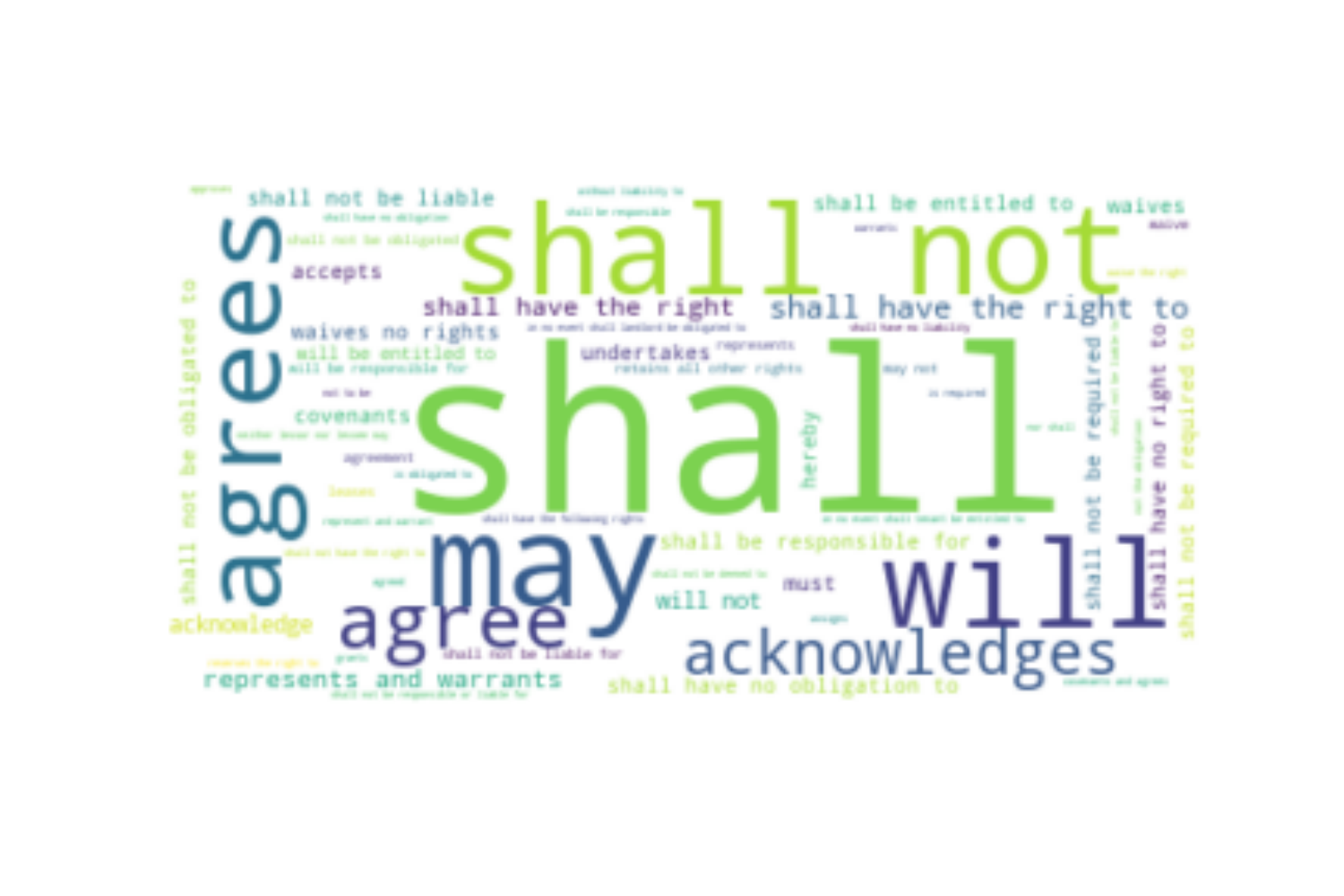}
    \caption{Frequency-based wordcloud of all the triggers.}
    \label{fig:trigger-wc}
\end{figure}

\noindent\textbf{Qualitative analysis.} Figure~\ref{fig:distribution} shows the distribution of annotated spans\footnote{For `None' type, no span is annotated so the bar denotes the number of sentences labeled as none.} over deontic types with respect to each agent (Landlord and Tenant for lease agreements\footnote{We add the statistics for tenant, subtenant \& lessee under Tenant, and landlord, sublandlord \& lessor under Landlord.}) in the train set. Interestingly, tenants have more obligations and prohibitions, and fewer entitlements or permissions than landlords. $17.3\%$ of the sentences have multiple trigger annotations, $48.6\%$ of these sentences express multiple deontic types. $24.8\%$ of the sentences do not express any deontic type with respect to a given agent. The dataset contains $383$ unique triggers across all the deontic types. Table~\ref{tab:topktrigger} lists the top $10$ triggers for each deontic type, and Figure~\ref{fig:trigger-wc} shows the frequency-based wordcloud of the annotated triggers. `Shall' constitutes $44.6\%$ of the annotated triggers used to express not only obligation but entitlement, no-obligation, and no-entitlement as well. Prohibitions may be expressed using negation words ($14.9\%$) between the context ({\it e.g.}, `neither lessor nor lessee may') of a sentence. While modal auxiliaries ({\it e.g.}, {\it shall}, {\it will}, {\it may}) are more frequently used, $45.2\%$ of the total unique triggers are non-modal expressions ({\it e.g.}, {\it agrees}, {\it represents}) covering $20.3\%$ of the annotated trigger spans. This shows that {\sc LexDeMod} covers a wide variety of linguistic expressions of deontic modality in legalese, not restricted to modal auxiliaries. Annotated samples from the dataset are provided in Table~\ref{tab:examples} in \ref{sec:example}.
\section{Proposed Benchmarking Tasks}
Having established the rich variety and coverage of linguistic expressions for deontic modality in {\sc LexDeMod}, we benchmark the corpus on the proposed two tasks defined below:\\
\noindent
    \textbf{(i) Agent-specific multi-label deontic modality classification. }This task aims at predicting all the deontic types expressed in a sentence with respect to an agent. We pose this as a multi-label classification task conditioned on a sentence and an agent. \\
    \noindent
    {\textbf{(ii) Agent-specific deontic modality and trigger span detection.} This task aims at identifying both the deontic type and corresponding triggers. We pose this as a token classification task. Every token in the corpus is assigned a BIOS tag if it belongs to a modal
trigger, which is appended with a suffix indicating its deontic type. For instance, Tenant$_{\sc O}$ \textit{is$_{\sc B-OBL}$ responsible$_{\sc I-OBL}$ for$_{\sc I-OBL}$} paying$_{\sc O}$ the$_{\sc O}$ rent$_{\sc O}$}, where subscripts denote the BIOS tags. 

For both the tasks, agent is conditioned using special tokens added at the beginning of a sentence. This simple strategy has been successfully used previously for controlled text generation tasks~\citep{sennrich2016controlling,johnson2017google,rudinger-etal-2020-thinking,Sancheti_Srinivasan_Rudinger_2022}.

\section{Benchmarking Setup}
We experiment with various pre-trained language models (PLMs)~\citep{devlin-etal-2019-bert,liu2019roberta}, which have shown state-of-the-art performance on natural language understanding tasks, to study their performance on our proposed tasks. We fine-tune these models for both the tasks on binary cross-entropy loss for $20$ epochs each with a batch size of $8$, and maximum sequence length of $256$ using HuggingFace's Transformers library~\citep{wolf-etal-2020-transformers}. The model(s) with the best macro-F1 score on the dev set is used to report results on the test set.
Further implementation details are in \ref{sec:implementation}.
We also partition the data according to the agent being conditioned to assess the performance of the trained models with respect to each agent.

\section{Benchmarking Multi-label Classification} \label{sec:eval-ml}
\noindent\textbf{Comparison models.}
We experiment with three kinds of approaches for the agent-specific multi-label deontic modality classification task.\\
\noindent \textbf{(1) Majority} class predicted for each agent.\\
\noindent \textbf{(2) Rule-based.} We implement a rule-based approach similar to the one described in~\citet{ash2020unsupervised} with additional conditioning on the agent. It searches for the presence of pre-defined modal triggers for a deontic type and associates it with the agent using dependency tags ({\it e.g.}, {\it nsubj}, {\it aux} or {\it agent}). We use spacy to tokenize each sentence and obtain a dependency parse. More details in~\ref{sec:rule-based}. \\
\noindent \textbf{(3) Fine-tuning PLMs.} We fine-tune the following PLMs differing in size and domain of data used for pre-training: \begin{inparaenum}[(i)]\item{BERT-base-uncased (\textbf{BERT-BU}); }\item{RoBERTa-base (\textbf{RoBERTa-B}); }\item{RoBERTa-large (\textbf{RoBERTa-L}), and }\item{recently introduced Contract-BERT-base-uncased (\textbf{C-BERT-BU}) model~\citep{chalkidis-etal-2020-legal} which has been pre-trained on US contracts from the EDGAR library.} \end{inparaenum}

All the above models are trained assuming trigger span information is not available and full context ({\it i.e.}, sentence) is used. To understand the importance of \textbf{A}gent conditioning, \textbf{C}ontext, and \textbf{T}rigger for this task, we additionally train the following models: 
\begin{inparaenum}[(i)]
    \item{\textbf{No-agent }where special token for agent is not used during training;}
    \item{\textbf{ACT-Masked }wherein everything in the context except the trigger span is masked using \texttt{[MASK]} token to hide the context but retain the positional information of the trigger; }
    \item{\textbf{AT }wherein only the tokens belonging to a trigger are used and multiple triggers are separated using a special token ({\it e.g.}, \texttt{[SEP]} or \texttt{</s>}), and }
    \item{\textbf{ACT }wherein all the triggers are appended at the end of the context separated by a special token ({\it e.g.}, \texttt{[SEP]} or \texttt{</s>}).}
\end{inparaenum}

\begin{table*}[t]
\centering
\scriptsize
\begin{tabular}{l c c c c}
\toprule
\textbf{Model} & \textbf{Accuracy} & \textbf{Precision} & \textbf{Recall} & \textbf{F1}\\
\midrule
Majority & $39.53/28.66/34.38$ &$6.49/5.23/11.72$ &$14.29/14.29/21.09$ &$8.92/7.66/15.03$  \\
Rule-based & $61.32/50.54/56.22$ &$\mathbf{81.81}/75.21/\mathbf{80.04}$ &$46.66/45.54/46.64$ &$50.13/46.16/48.76$\\
\midrule
BERT-BU & $74.07/70.79/72.52$ &$73.68/74.44/75.48$ &$75.84/71.02/77.17$ &$\mathbf{78.81}/71.18/75.61$ \\
RoBERTa-B & $75.53/71.42/73.59$ &$73.54/72.17/74.48$ &$78.39/72.88/78.31$ &$74.90/71.91/75.66$\\
C-BERT-BU & $77.50/73.25/75.48$ &$76.63/76.22/77.52$ &$\mathbf{80.47}/71.54/78.81$ &$77.95/72.34/77.67$ \\
RoBERTa-L & $\mathbf{78.28/75.03/76.74}$ &$75.05/\mathbf{77.69}/77.30$ &$79.59/\mathbf{75.21}/\mathbf{79.11}$ &$76.71/\mathbf{76.00}/\mathbf{77.88}$ \\
\midrule
RoBERTa-L-No-agent &$51.28/47.45/49.46$ &$57.09/53.79/58.32$ &$65.01/55.52/60.08$ &$55.75/52.56/57.53$ \\
RoBERTa-L-ACT-Masked & $81.52/72.02/77.02$ &$76.39/71.31/81.46$ &$71.22/65.74/75.90$ &$84.25/76.29/80.42$\\
RoBERTa-L-AT & $84.72/76.22/80.70$ &$79.84/73.60/82.29$ &$78.00/72.96/82.47$ &$87.58/80.58/84.20$ \\
RoBERTa-L-ACT & $\mathbf{91.66/88.62/90.23}$ &$\mathbf{88.44/85.40/89.48}$ &$\mathbf{88.10/84.43/89.21}$ &$\mathbf{93.38/91.38/92.42}$ \\
\bottomrule
\end{tabular} 
\caption{Evaluation results for agent-specific multi-label deontic modality classification task. Scores for \textbf{Tenant/Landlord/Both} are averaged over $3$ different seeds. BU, B, L, A, C, and T denote base-uncased, base, large, agent, context (sentence), and trigger, respectively. Development set results are provided in Table~\ref{tab:dev-ml} in Appendix.}\label{tab:ml-all}
\end{table*}
\noindent\textbf{Evaluation measures. }
We report macro-averaged Precision, Recall, and F1 scores across all the types, calculated using Sklearn library~\citep{pedregosa2011scikit}. We also report the Accuracy of predicting all the classes correctly for a sentence. 

\noindent\textbf{Results and analysis.} We report the results for multi-label classification task in Table~\ref{tab:ml-all}. While rule-based approach has better F1 score than majority type prediction for each agent, Transformer-based models outperform these baselines indicating their ability to better capture the linguistic diversity of expressing deontic modals. As expected, Rule-based approach has the highest overall precision but low recall due to the impossibility of enumerating all the rules. While C-BERT-BU, which is pre-trained on contracts, performs better than BERT-BU and RoBERTa-B, interestingly it achieves comparable F1 score to RoBERTa-L. This indicates that improvements from domain-specific pre-training may also be achieved with larger model size and more training data. 

As RoBERTa-L performs the best on this task, we report the results for variants of this model to understand the importance of agent conditioning, context, and trigger, in the last block of Table~\ref{tab:ml-all}. The performance of {RoBERTa-L-No-agent}, trained without agent conditioning, significantly drops as compared to RoBERTa-L, indicating the importance of agent conditioning during training and association of agent with the modality expressed in a sentence. Using trigger information during training ({RoBERTa-L-ACT}) significantly improves the performance over RoBERTa-L across all the measures, showing that triggers are indicative of specific deontic type. Higher scores for RoBERTa-L-AT than RoBERTa-L-ACT-Masked show that positional information of trigger span adds noise to the representations learned by the model. Further, context is also important for identifying deontic type, as all the metric scores drop when context is masked (RoBERTa-L-ACT-Masked) or not used (RoBERTa-L-AT) during training as compared to using all the information (RoBERTa-L-ACT).

Manual inspection of deontic type-wise (Table~\ref{tab:type-wise}) performance reveals that permission is the easiest, while no-entitlement and prohibition are the hardest to identify. This can be due to the use of limited variety in expressing permissions (majorly `may'), while use of negation within context for expressing prohibitions which makes it harder to identify. For tenant, obligation is identified more accurately than entitlements (vice-versa for landlord) as expected due to higher frequency of obligations for tenant and entitlements for landlord. 

Figure~\ref{fig:ml-per} shows the trend for RoBERTA-L's F1 score as train data size varies indicating that the rate of increase in F1 decreases with additional data.
\section{Benchmarking Trigger Span Detection} \label{sec:eval-bios}
\noindent\textbf{Comparison models.} We experiment with three kinds of approaches for the agent-specific deontic modality and trigger span detection task. \\
\noindent \textbf{(1) Majority. }`Shall' is the most used trigger as shown in Figure~\ref{fig:trigger-wc} and is used to express obligations for Tenant while entitlements for Landlord. This baseline tags each occurrence of `shall' with {\sc S-OBL} for tenant or {\sc S-ENT} for landlord as agent.\\ 
\noindent \textbf{(2) Rule-based. } We tag occurrences of pre-defined modal triggers in a sentence with the deontic type predicted using the rule-based approach (\textsection\ref{sec:eval-ml}).\\
\noindent \textbf{(3) Fine-tuning PLMs. }We fine-tune the same models as described in \textsection\ref{sec:eval-ml} on a token classification task to predict the BIOS tags. Additionally, we train a `No-agent' model to verify the importance of agent conditioning.
  
\begin{figure}[t]
    \centering
    \begin{subfigure}[t]{\columnwidth}
        \centering
    \includegraphics[width=0.99\columnwidth]{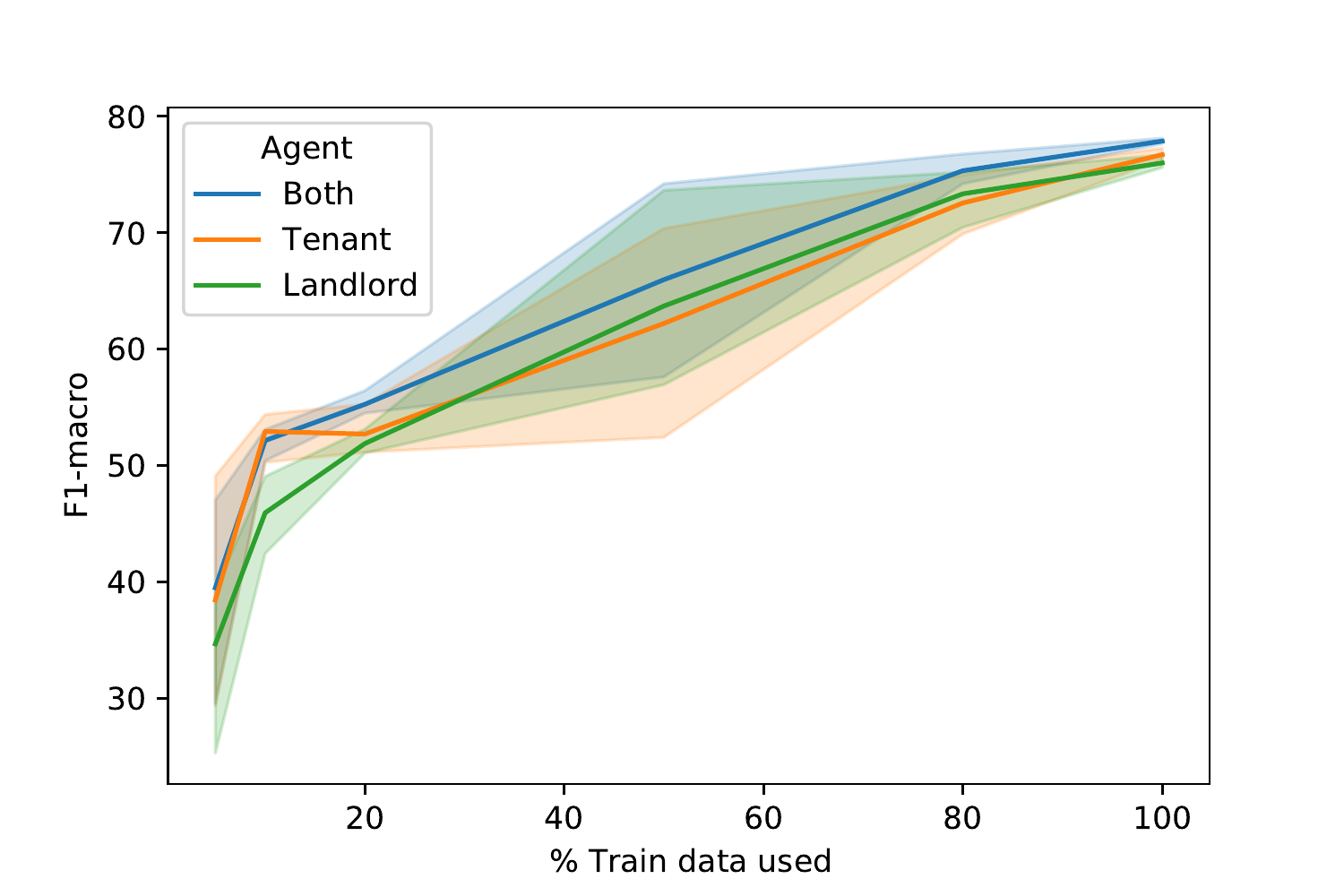}
        \caption{Classification} \label{fig:ml-per}
    \end{subfigure}
    \begin{subfigure}[t]{\columnwidth}
        \centering
        \includegraphics[width=0.99\columnwidth]{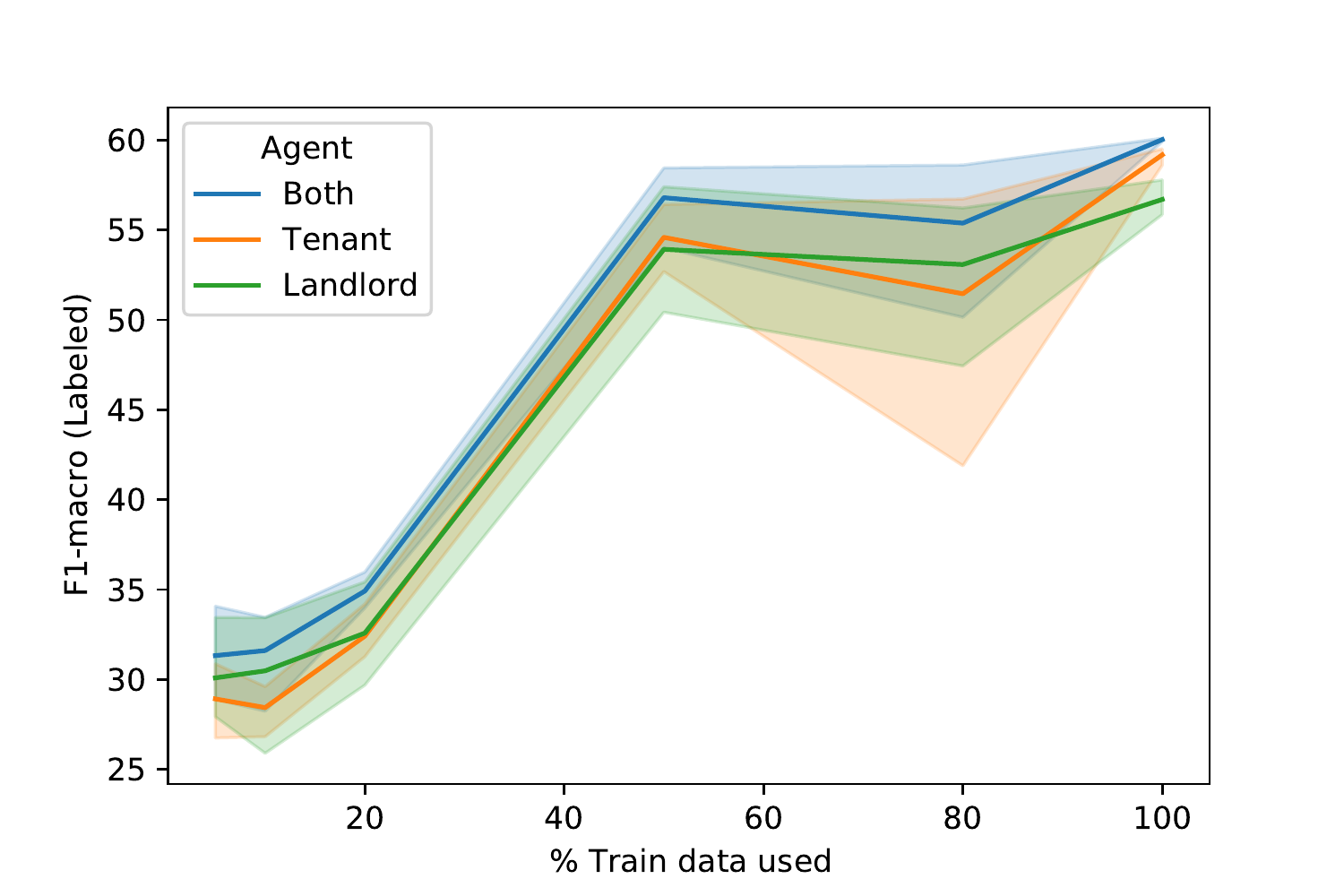}
    \caption{Trigger Span Detection} \label{fig:bios-per}
    \end{subfigure}
    \caption{RoBERTA-L's performance with varying train dataset size for the two tasks.} \label{fig:per-eval}
\end{figure}

\noindent\textbf{Evaluation measures. }
We report macro-averaged Precision, Recall, and F1 scores, calculated using seqeval library~\citep{seqeval}. We also report the Accuracy of predicting the BIOS tags for a sentence. Following~\citep{pyatkin2021possible}, we report these metrics in labeled (both deontic type and trigger span considered) and unlabeled (only trigger span without deontic type is considered) settings.
\begin{table*}[h]
\centering
\resizebox{0.99\textwidth}{!}{
\begin{tabular}{l c c c c| c c c c}
\toprule
\multirow{2}{*}{\textbf{Model}} & \multicolumn{4}{c|}{\textbf{Labeled}} & \multicolumn{4}{c}{\textbf{Unlabeled}} \\
 & \textbf{Accuracy} & \textbf{Precision} & \textbf{Recall} & \textbf{F1} & \textbf{Accuracy} & \textbf{Precision} & \textbf{Recall} & \textbf{F1} \\
\midrule
Majority & $97.16/96.85/97.01$ &$5.58/4.40/9.98$ &$10.89/10.31/16.00$ &$7.38/6.17/12.27$ &$97.28/97.04/97.17$ &$41.30/39.59/40.51$ &$50.61/42.86/46.76$ &$45.48/41.16/43.41$ \\
Rule-based & $97.85/97.67/97.76$ &$\mathbf{77.42/76.68/79.66}$ &$32.42/33.24/33.61$ &$40.00/39.30/40.58$ &$97.89/97.73/97.81$ &$\mathbf{72.59/73.97/73.22}$ &$40.07/35.34/37.73$ &$51.64/47.83/49.80$\\
\midrule
BERT-BU & $98.45/98.36/98.41$ &$53.04/56.11/56.48$ &$58.49/59.05/61.97$ &$55.11/\mathbf{57.01}/58.80$ &$\mathbf{98.55}/\mathbf{98.52}/\mathbf{98.53}$ &$68.87/69.92/69.38$ &$76.07/75.22/75.65$ &$72.29/72.46/72.38$\\
RoBERTa-B & $98.40/98.24/98.32$ &$53.03/52.43/55.57$ &$63.65/\mathbf{59.63/64.00}$ &$57.08/55.31/53.91$ &$98.49/98.41/98.46$ &$68.99/67.44/68.22$ &$78.18/\mathbf{75.36}/\mathbf{76.78}$ &$73.27/71.14/72.22$\\
C-BERT-BU & $98.44/\mathbf{98.39/98.42}$ &$53.46/54.70/57.08$ &$60.76/57.37/62.42$ &$56.45/55.68/59.31$ &$98.52/\mathbf{98.52}/98.52$ &$69.49/70.85/70.15$ &$76.89/74.26/75.59$ &$72.99/\mathbf{72.52}/72.76$\\
RoBERTa-L & $\mathbf{98.45}/98.27/98.37$ &$54.99/55.58/57.37$ &$\mathbf{65.55}/58.88/63.74$ &$\mathbf{59.19}/56.71/\mathbf{60.04}$ &$98.54/98.39/98.47$ &$69.78/69.56/69.68$ &$\mathbf{79.16}/74.35/\mathbf{76.78}$ &$\mathbf{74.18}/71.88/\mathbf{73.06}$\\
\midrule
RoBERTa-L-NA & $97.64/97.75/97.69$ &$32.45/36.71/36.36$ &$48.92/43.93/46.79$ &$34.68/38.72/39.45$ &$98.26/98.22/98.24$ &$61.73/64.63/63.12$ &$75.21/72.84/74.03$ &$67.79/68.46/68.11$\\
\bottomrule
\end{tabular}
}
\caption{Evaluation results for agent-specific modal trigger span detection task. Macro-averaged Precision, Recall and F1 scores are presented for \textbf{Tenant/Landlord/Both}. Scores are averaged over $3$ different seeds. BU, B, L, and NA denote base-uncased, base, large, and no-agent respectively. Dev set results are shown in Table~\ref{tab:dev-bios} in Appendix.}\label{tab:bios-all}
\end{table*}
\noindent\textbf{Results and Analysis. }Labeled and unlabeled metric scores for trigger span detection task are reported in Table~\ref{tab:bios-all}. RoBERTa-L has the best labeled F1 score which evaluates for both trigger detection and its correct deontic type identification. Similar to the classification task, Rule-based approach outperforms other models on precision however, lags behind in recall for the same reason. Size of the model (RoBERta-L) is instrumental than domain knowledge of C-BERT-BU. Consistently higher unlabeled scores, indicate that models are able to identify the trigger words. However, associating triggers with the correct deontic type is a harder task, owing to the multiple deontic types that a trigger can be used to express ({\it e.g.}, shall in Table~\ref{tab:topktrigger}) them. 
Similar to the classification task, importance of agent conditioning is evident from the last row with significant drop in F1 scores (even lower than Rule-based approach in Labeled score). Higher accuracy scores are due to the majority tokens being labeled as `O'. Trends with dataset size variation are shown in Figure~\ref{fig:bios-per}. Manual analysis of deontic type-wise span detection (Table~\ref{tab:type-wise}) reveals that prohibition, no-entitlement, and no-obligation are hard to identify. Similar trends were observed for tenant and landlord as in \textsection\ref{sec:eval-ml}. 

These results show that identification of both triggers and associating it with the deontic type is a difficult task owing to the linguistic variety of expressions used in legal language. 
\section{Beyond Lease Contracts} \label{sec:transfer}
To investigate if the diverse linguistic expressions used for expressing deontic modality is specific to a contract type, we collect annotations  via AMT using the same annotation protocol (\textsection\ref{sec:protocol}) for: \begin{inparaenum}[(1)]
\item{$470$ sentences from $3$ employment contracts in the LEDGAR corpus, and}
\item{$154$ sentences from $4$ rental agreement templates freely available at PandaDoc.\footnote{\url{https://www.pandadoc.com/} We chose rental agreements as they are commonly used by layperson than lease contracts from SEC.}}
\end{inparaenum}
 We evaluate the performance of the best model (RoBERTa-L) for both the tasks on these sentences and report the results in Table~\ref{tab:transfer}. We observe a performance drop (more for employment contracts than rental agreements) when compared to model's performance on lease agreements, although it is significantly better than the rule-based approach demonstrating the non-robustness of rule-based approach towards diverse linguistic expressions. This drop is more prominent for employment contracts due to the lease-specific agent conditioning ({\it e.g.}, tenant) used during training while commonly occurring agents in employment contracts are employee, employer, etc. 

To account for this, we additionally train models with anonymizing the agent mentions in the dataset in two ways: \begin{inparaenum}[(i)]\item{\textbf{RoBERTa-L-AR}-- all occurrences of an agent are replaced with the same token ({\it e.g.}, `a1' for tenant), and }\item{\textbf{RoBERTa-L-ARR} --agent is randomly replaced with a token consistent within a sentence. }\end{inparaenum} Replacing agent mentions leads to significant improvements for employment contracts in both the tasks, although evaluating these models on rental agreements (see Table~\ref{tab:transfer}) and lease data shows (see Table~\ref{tab:beyond-lease}) an expected drop in the performance. These experiments show that the linguistic expressions captured by \textsc{LexDeMod} are also generalizable to other types of contracts.
\begin{table}[t!]
\centering
\scriptsize
\resizebox{\columnwidth}{!}{
\begin{tabular}{c c c c c}
\toprule
{\textbf{Model}} & \textbf{Accuracy} & \textbf{Precision} & \textbf{Recall} & \textbf{F1}\\
\midrule
\multicolumn{5}{c}{\textbf{Multi-label Classification (Rental/Employment)}}\\
\midrule
Majority & $36.36/27.45$ & $11.87/8.80$ & $19.10/15.15$ & $14.46/11.11$\\
Rule-based & $41.56/47.45$ & $53.77/64.63$ & $34.54/35.00$ & $33.27/37.22$\\
\midrule
RoBERTa-L & $73.16/48.72$ & $83.08/52.87$ & $63.42/48.90$ & $68.90/48.32$\\
\midrule
RoBERTa-L-AR &  $55.19/42.55$ & $56.87/59.29$ & $52.38/46.48$ & $50.66/50.30$\\
RoBERTa-L-ARR & $70.35/64.68$ & $76.79/70.05$ &	$63.14/64.62$ & $65.89/65.36$\\
\midrule
\multicolumn{5}{c}{\textbf{Trigger Span Detection (Labeled) (Rental/Employment)}}\\
\midrule
Majority & $96.09/97.37$ & $18.33/4.23$ & $1.90/7.08$ & $3.42/5.30$\\
Rule-based & $96.40/97.83$ & $56.25/59.66$ & $23.69/19.65$ & $29.62/27.45$\\
\midrule
RoBERTa-L & $97.48/97.78$ & $49.74/36.80$ & $45.87/37.84$ & $45.58/34.87$\\
\midrule
RoBERTa-L-AR & $97.22/98.15$ & $49.97/48.86$ & $44.43/42.99$ & $44.22/43.42$\\
RoBERTa-L-ARR & $97.60/98.38$ & $59.42/53.14$ & $47.83/43.84$ & $49.61/45.47$\\
\bottomrule
\end{tabular}
}
\caption{Results for rental/employment contracts.\protect\footnotemark}\label{tab:transfer}
\end{table}
\footnotetext{Unlabeled scores are provided in Appendix in Table~\ref{tab:transfer-un}.}
\section{Case Study: Red flag Detection} \label{sec:redflag}
To investigate if our agent-specific deontic modality classifier is capable of identifying the red flags annotated by~\citet{leivaditi2020benchmark} for lease agreements, we compare the predictions on the dev set from ALeaseBERT, proposed by~\citet{leivaditi2020benchmark}, and RoBERTa-L model trained on \textsc{LexDeMod} dataset. For each sentence in the red flags dataset, we predict the deontic modality with respect to each of the agent alias mentioned in that sentence. If any one of the deontic types is expressed for any of the agents then we consider the prediction as positive otherwise negative. We find that (see Table~\ref{tab:case} in \ref{sec:app-redflag}) the model trained on \textsc{LexDeMod} has high recall and low precision while ALeaseBERT has high precision but low recall for the positive class. Our model was able to predict all the red flags predicted by ALeaseBERT and some additional red flags. This is expected as many permissions or entitlements may not be red flags but may belong to a deontic type. We also found that there were payments related obligations which were predicted as red flags by our model but were not annotated as red flags in the dataset. Therefore, our model could also be used to filter important sentences which could indicate some red flags due to high recall.

\section{Conclusion and Future Work}
We introduced {\sc LexDeMod} for deontic modality detection in the legal domain which consists of diverse linguistic expressions of deontic modality. We propose and benchmark two tasks namely, agent-specific multi-label deontic modality classification, and agent-specific deontic modality and trigger span detection using transformer-based models. 
While evaluation results are promising, there is substantial room for improvement. We demonstrated the generalizability of the diverse linguistic expressions captured in {\sc LexDeMod} via transfer learning experiments to employment and rental lease agreements. The small case study on red flag detection using our data showed the usability of our dataset. 
We leave joint-modeling of the two tasks and using these identification models for generating ``at a glance'' summary of contracts for future work. 
\section{Limitations}
We note a few limitations: \begin{inparaenum}[(1)]
\item{Although we demonstrate reasonable generalization to employment agreements, our dataset is limited to lease agreements which may not cover all the linguistic expressions for deontic modality occurring in legal domain.}
\item{The custom interface built for collecting annotations does not support non-contiguous trigger-span selection which may result in some contract type specific triggers (only for triggers with negation). Future work may consider handling non-contiguous spans and other challenges associated with it (\textit{e.g.}, representing non-contiguous trigger spans for a category in the BIO span).}
\item{As we focus on identifying agent-specific deontic modalities, we only consider sentences where the agent alias is explicitly mentioned. This helped in simplifying the annotation process and cost efficiency. Therefore, our models may not work well when no agent alias is mentioned in the given sentence. We leave the collection of annotations for sentences not explicitly mentioning agent alias for future work.}
\item{Our data collection and modeling assume that agent alias is known apriori (for which we perform agent alias extraction) as we focus on the identification task. Extending this work to any other type of agreement will require similar alias extraction method (\textit{e.g.}, employee, employer for employment agreement) or a more sophisticated model to identify the agent implicitly.}
\end{inparaenum}
\section{Ethical Considerations}
We are committed to ethical practices and protecting the anonymity and privacy of the annotators who have contributed. We paid annotators at an hourly rate of $7.5$ USD for their annotations.

\noindent \textbf{Societal Impact. }We recognize and acknowledge that our work carries a possibility of misuse including malicious adulteration of summaries generated by extracting sentences identified by our model and adversarial use of the model to mislead users. Such kind of misuse is common to any prediction model therefore, we strongly recommend coupling any such technology with external expert validation. The purpose of this work is to provide aid to the legal personnel or layperson dealing with legal contracts for a better understanding of the legal documents, and not to replace any experts. As contracts are long documents, identification of sentences that express deontic types can help in significantly reducing the number of sentences to read or highlighting the important parts of the contract which may need more attention.
\section{Acknowledgements}
We would like to thank Ani Nenkova and the anonymous reviewers for their useful feedback and comments on this work. We acknowledge the support from Adobe Research unrestricted gift funding for this work. The views contained in this article are those of the authors and not of the funding agency.
\bibliography{anthology,custom}

\clearpage
\appendix

\section{Appendix} \label{sec:appendix}
\subsection{Combining Bullets} \label{sec:combine-bullet}
We combine the higher level context (bullets- ``parent'') with the lower level context (sub-bullet- ``child'') owing to the hierarchical nature of contracts by iterating over the provisions in a contract in sequential order and following the below rules. Combination can be done in two ways: \begin{inparaenum}[(i)]\item{concatenating, and }\item{merging. }\end{inparaenum} We find a sub-bullet via regular expression (\verb|^\([ivx]+||\verb|^\([a-zA-Z]+| |\verb|^[\d.\d]+| ) pattern matching.
\begin{itemize}
    \item If the child is not a complete sentence (identified by the presence of S in root of constituency parse), parent is a complete sentence, and parent does not contain `follow'  or `below:' then remove `:' from the end of parent and append the child (we call this, merging).
\item If child starts with a lower case and parent does not contain `follow' or `below:' then remove `:' and append child irrespective of the root label of constituency.
\item If parent ends with `the following:' then remove `the following:' and append the child if it is not a complete sentence else do not remove `the following:' and just append the child (we call this, concatenating).
\item If none of the above rule satisfies and the parent ends with a `:' then just concatenate the child with the parent.
\end{itemize}

\subsection{Annotation Guidelines} \label{sec:ann-guide}
\begin{figure*}[t!]
    \centering
    \includegraphics[width=0.7\textwidth,height=8cm]{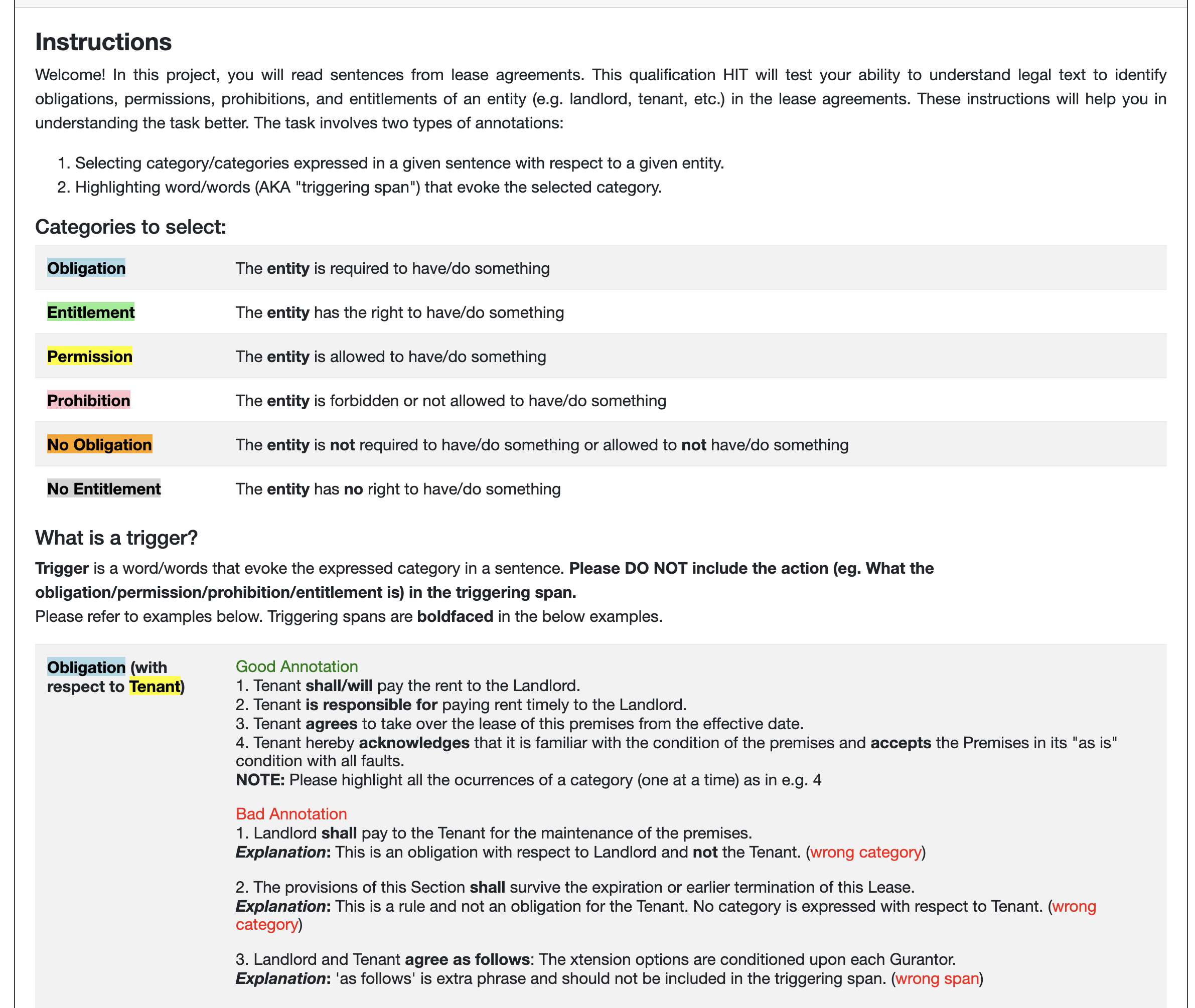}
    \includegraphics[width=0.7\textwidth,height=8cm]{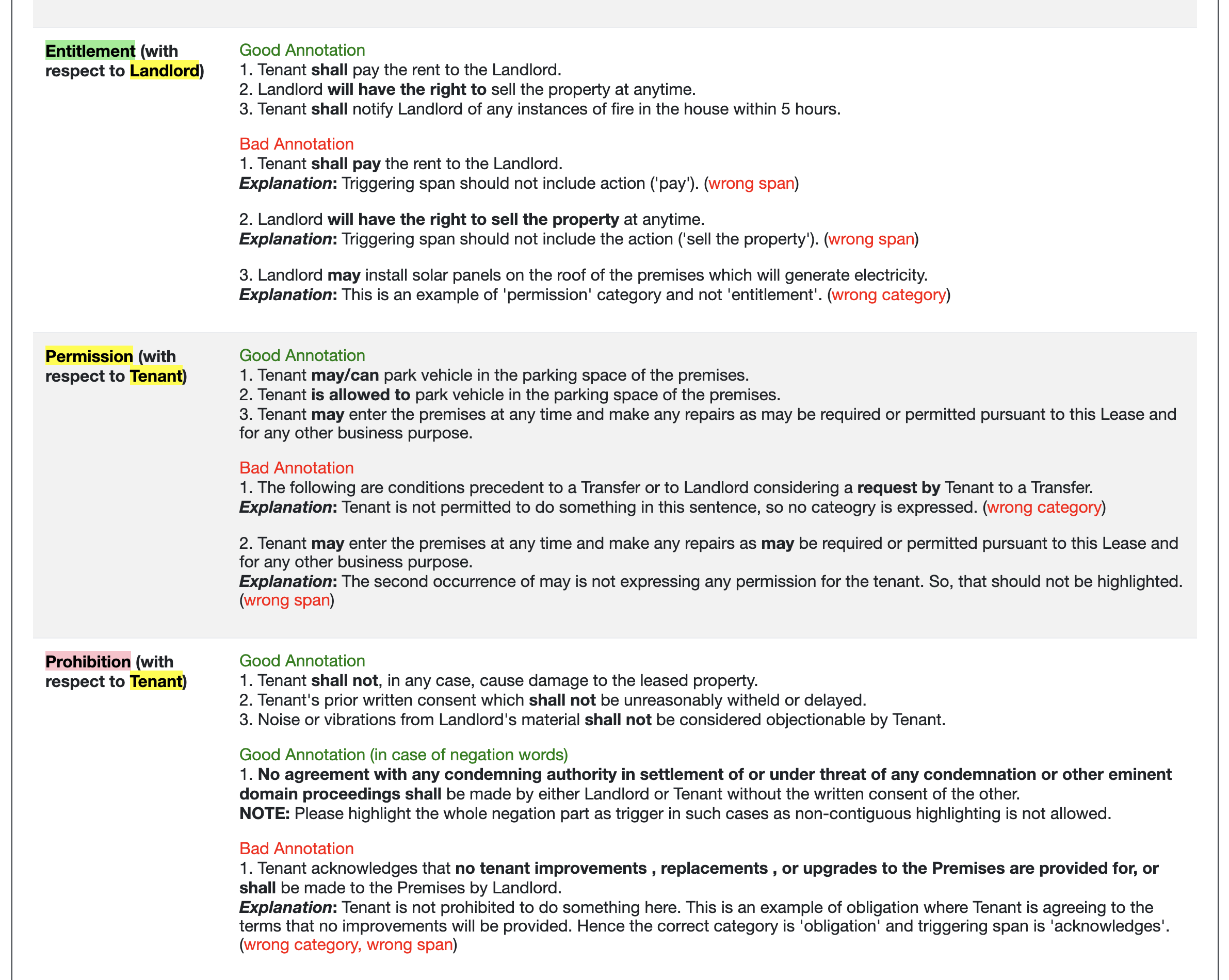}
    \includegraphics[width=0.7\textwidth,height=8cm]{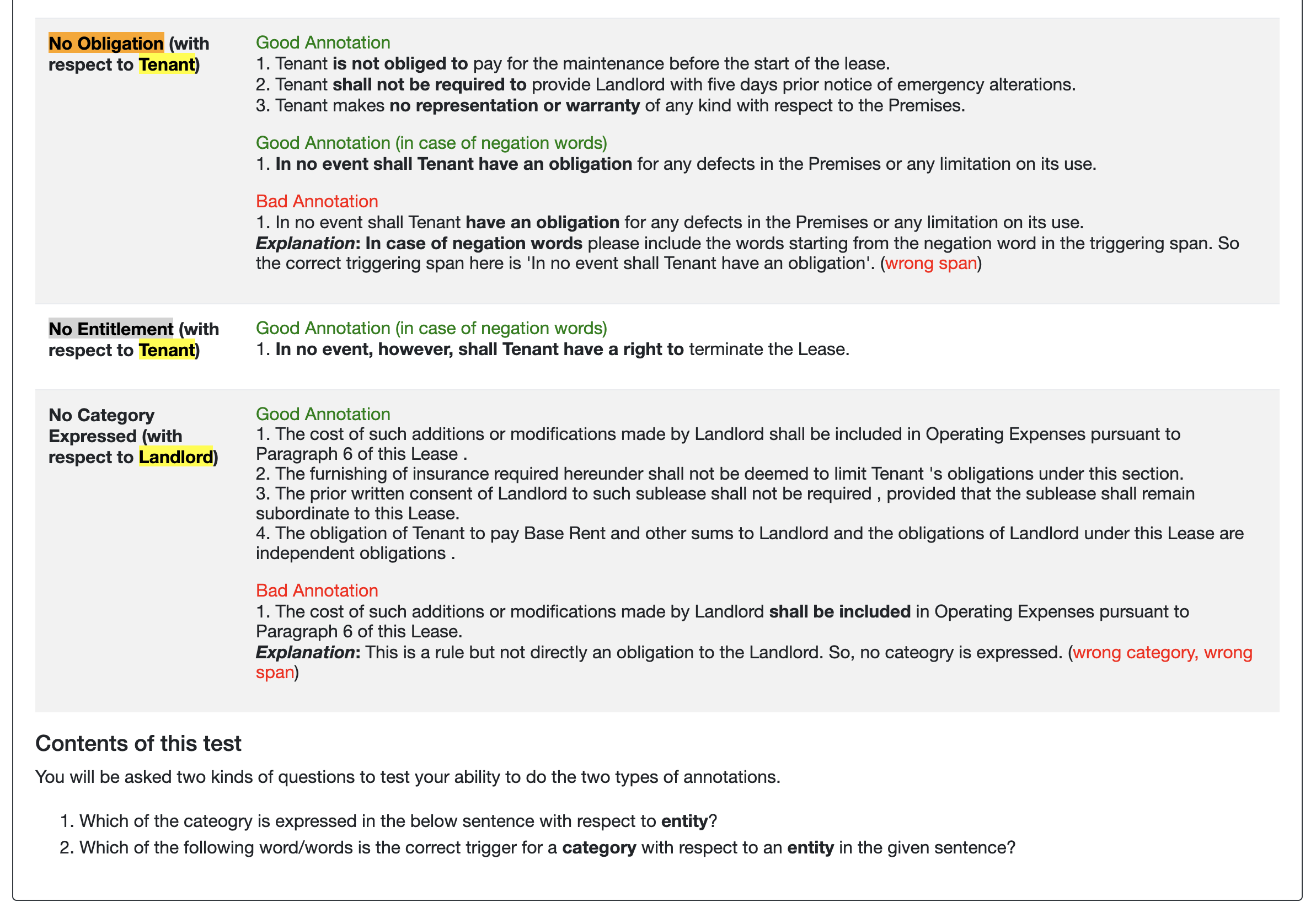}
    \caption{Instructions and examples provided to the annotators.}
    \label{fig:guidelines}
\end{figure*}

We present the instructions, and the correctly and incorrectly annotated examples with explanations provided to the annotators in Figure~\ref{fig:guidelines}. The custom annotation inference built to collect the data is shown in Figure~\ref{fig:interface}. We manually annotate $~50$ sentences and use them as quality check questions to ensure annotators are sincerely and correctly annotating each HIT. Type-wise inter-annotator agreement for the sentences in test split is shown in Table~\ref{tab:agreement}.
\begin{figure*}[h!]
    \centering
    \includegraphics[width=0.7\textwidth,height=8cm]{figs/interface.pdf}
    \caption{Annotation Interface.}
    \label{fig:interface}
\end{figure*}
\begin{table}[t!]
\centering
\resizebox{0.99\columnwidth}{!}{
\begin{tabular}{c c c c c c c}
\toprule
\textbf{Obl} & \textbf{Ent} & \textbf{Pro} &\textbf{Per} & \textbf{Nobl} & \textbf{Nent} & \textbf{None}\\
\midrule
$0.82$& $0.68$ &$0.44$ &$0.82$& $0.76$ & $0.41$ & $0.65$\\
\bottomrule
\end{tabular}
}
\caption{Deontic type-wise inter-annotator agreement ($\alpha$) for the test set.}\label{tab:agreement}
\end{table}

\subsection{Qualification Questions} \label{sec:qualification}
We ask $10$ multiple choice questions in the pre-qualification task consisting of $5$ questions to test the understanding of identifying the correct deontic type and $5$ questions to test their understanding of trigger span selection for a deontic type.
\subsection{Resolving Disagreements} \label{sec:diagreements}
Disagreement in the annotation for duplicate sentences is resolved by one of the authors. The disagreement could occur because of any missing modality in case of multiple modalities expressed in a sentence, incorrect interpretation of the sentence, or human error in terms of annotating with respect to a tenant or a landlord. Consider the below sentence: ``[landlord] After final approval of the Final Plans by applicable governmental authorities, no further changes may be made thereto without the prior written approval of both Landlord and Tenant.", it was annotated as `prohibition' for landlord by one of the annotators and `none' by another annotator. As the prohibition mentioned in the sentence is not for the landlord, the correct annotation is `none'. Therefore, we retain the correct annotation for the example and discard the sentence with the incorrect annotation. Another example is ``[landlord] All conditions and agreements under the Lease to be satisfied or performed by Landlord have been satisfied and performed." which was incorrectly annotated as an `obligation'.
\begin{table}[t!]
\centering
\scriptsize
\resizebox{0.99\columnwidth}{!}{
\begin{tabular}{l |p{5cm}}
\toprule
\textbf{Type} & \multicolumn{1}{c}{\textbf{Heuristic triggers}}\\
\midrule
Obl &  shall/will be required, shall be obligated, shall, must, will, have to, should, ought to have, will/shall be paid\\
Ent & shall/will be entitled, shall/will be paid, shall/will retain, shall/will receive, shall have the right to, shall be retained, shall be kept, shall be claimed, shall be accessible, shall be owned, shall be determined\\
Pro & shall/will/must/may not, cannot, shall have no right, can not, shall/will not be allowed, shall not assist, shall/will be prohibited\\
Per & shall be permitted, shall also be permitted, can, may, could, shall/will be allowed \\
Nobl & shall/will not be liable for, shall/will not be obligated to, shall/will not be obligated for, shall/will not be responsible for, shall/will not be required to\\
Nent & shall/will not entitled to, shall/will not have the right to,  shall/will not be entitled for\\
\bottomrule
\end{tabular}
}
\caption{Triggers used to identify the deontic types.}\label{tab:rules}
\end{table}
\begin{table*}[h!]
\centering
\scriptsize
\begin{tabular}{l c c c c}
\toprule
\textbf{Model} & \textbf{Accuracy} & \textbf{Precision} & \textbf{Recall} & \textbf{F1}\\
\midrule
Majority &  $47.87/26.06/38.48$ &$7.60/4.83/12.43$ &$14.29/14.29/20.29$ &$9.92/7.22/15.26$\\
Rule-based & $52.13/42.25/47.88$ &$63.81/65.28/75.54$ &$48.08/32.69/40.07$ &$47.42/34.63/42.75$\\
\midrule
BERT-BU & $75.71/67.61/72.22$ &$73.80/68.20/72.50$ &$76.14/61.52/72.56$ &$74.16/62.42/72.13$ \\
RoBERTa-B & $72.69/69.48/71.31$ &$73.94/74.61/73.71$ &$76.06/74.94/76.50$ &$73.65/73.86/74.36$ \\
RoBERTa-L & $77.13/67.37/72.93$ &$76.93/68.95/74.29$ &$78.01/69.14/75.73$ &$76.79/67.76/74.32$\\
C-BERT-BU & $76.60/68.31/73.03$ &$78.17/69.27/75.52$ &$81.19/68.62/77.43$ &$79.14/67.49/75.92$\\
\bottomrule
\end{tabular} 
\caption{Evaluation results for agent-specific multi-label deontic modality classification task on development set. Scores are averaged over $3$ different seeds. BU, B, and L denote base-uncased, base, and large respectively.}\label{tab:dev-ml}
\end{table*}
\begin{table*}[h!]
\centering
\resizebox{0.99\textwidth}{!}{
\begin{tabular}{l c c c c| c c c c}
\toprule
\multirow{2}{*}{\textbf{Model}} & \multicolumn{4}{c|}{\textbf{Labeled}} & \multicolumn{4}{c}{\textbf{Unlabeled}} \\
 & \textbf{Accuracy} & \textbf{Precision} & \textbf{Recall} & \textbf{F1} & \textbf{Accuracy} & \textbf{Precision} & \textbf{Recall} & \textbf{F1} \\
\midrule
Majority & $97.26/96.93/97.11$ &$7.23/4.73/11.96$ &$10.43/10.42/15.52$ &$8.54/6.50/13.36$ &$97.41/97.20/97.32$ &$52.91/46.81/50.30$ &$51.81/44.00/48.40$ &$52.36/45.36/49.33$\\
Rule-based & $97.72/97.57/97.65$ &$58.27/58.99/72.43$ &$35.61/17.53/25.83$ &$40.13/25.52/34.36$ &$97.76/97.62/97.70$ &$80.22/66.67/75.35$ &$37.82/22.67/31.20$ &$51.41/33.83/44.12$\\
\midrule
BERT-BU & $98.23/98.03/98.14$ &$57.07/45.78/54.75$ &$57.81/45.74/55.23$ &$56.91/44.18/54.14$ &$98.41/98.29/98.36$ &$72.70/67.68/70.47$ &$72.37/67.33/70.17$ &$72.49/67.50/70.30$\\
C-BERT-BU & $98.18/97.95/98.08$ &$54.09/60.30/56.96$ &$55.31/53.18/57.01$ &$54.12/52.86/55.73$ &$98.35/98.20/98.29$ &$69.58/68.08/68.90$ &$70.99/68.67/69.97$ &$70.26/68.37/69.43$\\
RoBERTa-B & $98.12/97.83/97.99$ &$55.33/48.38/53.63$ &$58.95/52.51/57.78$ &$56.73/49.80/55.31$ &$98.28/98.06/98.18$ &$70.30/67.56/69.08$ &$73.40/71.78/72.69$ &$71.80/69.56/70.81$\\
RoBERTa-L & $98.09/97.83/97.97$ &$56.81/47.91/54.88$ &$60.63/51.78/59.10$ &$57.69/49.29/56.33$ &$98.23/98.05/98.15$ &$70.67/66.11/68.64$ &$73.58/69.56/71.82$ &$72.09/67.78/70.19$\\
\midrule
RoBERTa-L-NA & $97.57/97.76/97.65$ &$35.59/44.35/40.60$ &$46.54/48.02/48.67$ &$36.81/45.7/43.16$ &$98.27/98.19/98.23$ &$69.75/69.00/69.43$ &$74.61/72.89/73.86$ &$72.08/70.86/71.55$\\
\bottomrule
\end{tabular}
}
\caption{Evaluation results for agent-specific modal trigger span detection task on development set. Macro-averaged scores for Tenant/Landlord/All are presented for precision, recall and F1 measures. Scores are averaged over $3$ different seeds. BU, B, and L denote base-uncased, base, and large respectively.}\label{tab:dev-bios}
\end{table*}
\subsection{Rule-based Approach} \label{sec:rule-based}
We first curate a pre-defined list of triggers (Table~\ref{tab:rules}) used to express deontic types in legal domain following~\citet{ash2020unsupervised}. Then, tokenize and obtain the dependency parse and part of speech (POS) tags each each token in a sentence using spaCy python library. We describe the heuristic algorithm (by observing patterns in the train set) which searches for the presence of pre-defined triggers in a given sentence to extract its position (start index), each of the agents' mention, and its dependency tag for a sentence in Algorithm~\ref{alg:rule-based}.
\begin{algorithm*}[t!]
\small
  \caption{\texttt{Rule-based Heuristic}}
  \label{alg:rule-based}
  \begin{algorithmic}[1]
    \State \textbf{Inputs:} List $T$ of pre-defined triggers,  List $A$ of aliases for the type of contract to process.  
     \State \textbf{Outputs:} List $L$ of tuples containing (Deontic type, trigger, agent, start index) for all the sentence in the contract.
    \State ${L} \gets [ ]$, ${I} \gets [ ]$   \hfill // Initialization
    \For{each sentence in contract}
        \State // Initialize a list to keep account of visited trigger indices
        \State ${visited} \gets [ ]$ 
        \For{each $t$ in $T$}
            \If{$t$ in sentence}
                \State // Initialize a list of trigger indices
                \State ${indices} \gets [ ]$
                \For{each $t$ in sentence}
                    \If{start index of $t \notin$ visited}
                    \State ${indices} \gets \text{start index}$
                    \State ${visited} \gets \text{start index}$
                    \EndIf
                \EndFor
                \For{word in sentence}
                    \If{word.dependency is ROOT or word.pos $\in$ [VERB, AUX]}
                        \For{child in word.children} \hfill // Iterate over the children of word in the dependency tree
                        \State If $a1\in A$ is `nsubj/nsubjpass' of word \& child==$t$[0] \& child.dependency is `aux' \& child.index in indices then $L \gets (Type(t), t, a1, child.index)$ \hfill // Rule 1
                        \State If Rule 1 \& $a2\in A$ is a `conj' of $a1$ then $L \gets (Type(t), t, a2, child.index)$ // Rule 2
                        \State If child1.dependency is `agent' \& child2==$t$[0] \& child2.dependency is `aux' \& $a1\in A$ in children(child1)=child3 then $L \gets (Type(t), t, a1, child2.index)$ \hfill // Rule 3 
                        \State If Rule 3 \& $a2\in A$ in conjunction of child3 \& VERB in conjunction of word \& $t1$ is `aux' of VERB then $L \gets (Type(t1), t1, a2, t1.index)$ \hfill // Rule 4
                        \State If Rule 3 \& not Rule 4 \& $a2\in A$ in conjunction of child3 \& child==$t$[0] \& child.dependency is `aux' \& child.index in indices then $L \gets (Type(t), t, a2, child.index)$ \hfill // Rule 5
                        \State If child.dependency in ['pobj', 'dobj'] \& $a1\in A$ is in conjunction of children(child)=child1 \& VERB in conjunction of word \& $t1$ is `aux' of VERB then $L \gets (Type(t1), t1, a1, t1.index)$ \hfill // Rule 6
                        \State If child==$t$[0] \& child.dependency is `aux' \& child.index in indices \& VERB in conjunction of word \& $t1$ is `aux' of VERB \& `agent' in children(conjunction VERB)= child1.dependency \& $a2\in A$ in children(child1) then $L \gets (Type(t), t, a2, child.index)$ \hfill // Rule 7 
                        \State If child==$t$[0] \& child.dependency is `aux' \& child.index in indices \& VERB in conjunction of word \& $t1$ is `aux' of VERB \& not Rule 7 then $L \gets (Type(t1), t1, Agent(t), t1.index)$ \hfill // Rule 8
                       \EndFor
                    \EndIf
                \EndFor
            \EndIf
        \EndFor  
     \EndFor
  \end{algorithmic}
\end{algorithm*}
\begin{table}[t!]
\centering
\resizebox{0.99\columnwidth}{!}{
\begin{tabular}{c c c c| c c c}
\toprule
{\textbf{Deontic}} & \multicolumn{3}{c|}{\textbf{Classification}} & \multicolumn{3}{c}{\textbf{Span Detection}} \\
{\textbf{Type}} & \textbf{Precision} & \textbf{Recall} & \textbf{F1} &  \textbf{Precision} & \textbf{Recall} & \textbf{F1} \\
\midrule
Obl &  $84.87$ &   $87.83$   & $86.32$ & $76.93$   & $80.80$ &   $78.82$   \\
Ent &    $79.20$  &  $85.65$ &   $82.30$ & $66.96$  &  $77.89$   & $72.02$\\
Pro &     $60.76$  &  $75.00$  & $67.13$  &   $48.94$ &   $68.66$ &    $57.14$ \\
Per &    $91.25$ &   $87.43$  &  $89.30$  & $90.79$&  $82.63$ &   $86.52$\\
Nobl &    $74.58$  &  $87.13$   & $80.37$ & $31.88$&    $38.94$   & $35.06$  \\
Nent &    $67.95$  &  $60.23$   & $63.86$  & $29.73$  &  $32.67$  &  $31.13$\\
None &    $82.60$&    $73.10$   & $77.56$  & $-$  & $-$  &  $-$  \\
\bottomrule
\end{tabular}
}
\caption{Deontic type-wise results for agent-specific multi-label classification and modal trigger span detection (labeled) task on test set from the best (out of $3$ seeds) RoBERTa-L model.}\label{tab:type-wise}
\end{table}
\begin{table}[t!]
\centering
\scriptsize
\begin{tabular}{c c c c c}
\toprule
{\textbf{Model}} & \textbf{Accuracy} & \textbf{Precision} & \textbf{Recall} & \textbf{F1}\\
\midrule
\multicolumn{5}{c}{\textbf{Multi-label Classification}}\\
\midrule
RoBERTa-L & $76.74$ & $77.30$ & $79.11$ & $77.88$\\
\midrule
RoBERTa-L-AR &  $47.91$ & $56.99$ & $59.91$ & $56.60$\\
RoBERTa-L-ARR & $66.18$	& $69.75$&	$74.14$	& $71.22$\\
\midrule
\multicolumn{5}{c}{\textbf{Trigger Span Detection (Labeled)}}\\
\midrule
RoBERTa-L & $98.37$ & $57.37$ & $63.74$ & $60.04$\\
\midrule
RoBERTa-L-AR & $98.12$ & $50.11$ & $59.54$ & $53.63$\\
RoBERTa-L-ARR & $98.42$ & $58.42$ & $64.88$ & $61.19$\\
\midrule
\multicolumn{5}{c}{\textbf{Trigger Span Detection (Unlabeled)}}\\
\midrule
RoBERTa-L & $98.47$ & $69.68$ & $76.78$ & $73.06$\\
\midrule
RoBERTa-L-AR & $98.36$ & $67.03$ & $75.76$ & $71.07$\\
RoBERTa-L-ARR & $98.54$ & $70.65$ & $76.76$ & $73.58$\\
\bottomrule
\end{tabular}
\caption{Evaluating RoBERTa-L-AR and RoBERTa-L-ARR on lease test set.}\label{tab:beyond-lease}
\end{table}
\begin{table}[t!]
\centering
\scriptsize
\resizebox{\columnwidth}{!}{
\begin{tabular}{c c c c c}
\toprule
{\textbf{Model}} & \textbf{Accuracy} & \textbf{Precision} & \textbf{Recall} & \textbf{F1}\\
\midrule
Majority & $96.09/97.53$ & $54.55/41.26$ & $4.55/34.16$ & $8.39/37.38$\\
Rule-based & $96.40/97.85$ & $84.62/73.28$ & $16.67/21.72$ & $27.85/33.51$\\
\midrule
RoBERTa-L & $97.78/98.09$ & $69.85/54.32$ & $68.44/60.76$ & $69.14/57.36$\\
\midrule
RoBERTa-L-AR & $97.81/98.31$ & $68.77/59.64$ & $69.44/57.05$ & $69.09/58.30$\\
RoBERTa-L-ARR & $97.78/98.46$ & $69.51/61.81$ & $70.45/57.50$ & $69.94/59.58$\\
\bottomrule
\end{tabular}
}
\caption{Unlabeled Metric scores for trigger span detection task on rental/employment contracts.}\label{tab:transfer-un}
\end{table}\subsection{Implementation Details} \label{sec:implementation}
We run each model on $3$ seed values. We use Adam optimizer with a linear scheduler for learning rate with an initial learning rate of $2e^{-5}$, and warm-up ratio set at $0.05$. All the models are trained and tested on NVIDIA Tesla V$100$ SXM$2$ $16$GB GPU machine. We experiment with batch size $\in \{2,4,8\}$, number of epochs $\in \{3,5,10,20,30\}$, learning rate $\in \{1e^{-5},2e^{-5},3e^{-5}, 5e^{-5}\}$, and warm-up ratio $\in \{0.05, 0.10\}$. BERT-base ($110$M parameters) and Roberta-base ($125$M parameters) models took $~46$mins, and RoBERTa-large ($355$M parameters) took $~2$hrs to train for each of the tasks.
\subsection{Additional Results}
Table~\ref{tab:dev-ml}, and~\ref{tab:dev-bios} shows the results for the two tasks on the dev set. Table~\ref{tab:type-wise} shows the type-wise results from the best performing model. Table~\ref{tab:beyond-lease} shows the performance of models trained with anonymized agent on the test set of lease contracts. Table~\ref{tab:transfer-un} shows the unlabeled metric scores for generalizability to rental and employment contracts.
\begin{table}[h!]
\centering
\begin{tabular}{c c c c}
\toprule
{\textbf{Model}} & \textbf{Precision} & \textbf{Recall} & \textbf{F1}\\
\midrule
ALeaseBERT&$82.35$&$8.09$&$14.74$\\
Ours&$8.53$&$87.28$&$15.54$\\
\bottomrule
\end{tabular}
\caption{Results from the red flag detection case study. Our (ALeaseBERT) denotes RoBERTa-L model trained on \textsc{LexDeMod} (Red flags dataset~\citep{leivaditi2020benchmark}).}\label{tab:case}
\end{table}
\begin{table*}[b!]
\centering
\scriptsize
\resizebox{0.99\textwidth}{!}{
\begin{tabular}{p{0.5cm}|p{14cm}}
\toprule
\textbf{Type} & \multicolumn{1}{c}{\textbf{Examples}}\\
\midrule
\multirow{2}{*}{\parbox{0.5cm}{\centering{Obl}}} & [tenant] Tenant \textbf{shall} repair any damage resulting from such removal and \textbf{shall} restore the Property to good order and condition. \\
& [tenant] Tenant \textbf{acknowledges} and \textbf{agrees} that Landlord shall have the right to adopt reasonable rules and regulations for the use and/or occupancy of the Leased Premises and Tenant \textbf{agrees} that it shall at all times observe and comply with such rules and regulations.\\
\hline
\multirow{3}{*}{\parbox{0.5cm}{\centering{Ent}}} &[tenant] Tenant \textbf{shall also have the right to} use the roof riser space of the Building.\\
&[lessor] Rent \textbf{shall} be payable at Lessor 's place of business , or such other place as Lessor may direct from time to time.\\
& [landlord] Landlord \textbf{reserves the right to} modify Common Areas, provided that such modifications do not materially adversely affect Tenant's access to or use of the Premises for the Permitted Use.\\
\hline
\multirow{2}{*}{\parbox{0.5cm}{\centering{Pro}}} &  [lessee] Lessee \textbf{shall not} commit or allow waste to be committed on the Premises, and Lessee \textbf{shall not} allow any hazardous activity to be engaged in upon the Premises.\\
& [lessor] \textbf{Neither Lessor nor Lessee may} record this Lease nor a short - form memorandum thereof.\\
\hline
\multirow{2}{*}{\parbox{0.5cm}{\centering{Per}}} & [tenant] Tenant \textbf{may}, without Landlord's consent, before delinquency occurs, contest any such taxes related to the Personal Property. \\
& [lessor] Additional keys \textbf{may} be furnished at a charge by Lessor.\\
\hline
\multirow{2}{*}{\parbox{0.5cm}{\centering{Nobl}}} & [tenant] For the avoidance of doubt, to the extent there is a bank vault in the Premises, Tenant \textbf{shall have no obligation to} remove such vault on surrendering the Premises.\\
& [lessor] Further, \textbf{in no event shall Lessor have any obligation to} repair any damage to, or replace any of Lessee's furniture, trade fixtures, equipment or other personal property.\\
\hline
\multirow{2}{*}{\parbox{0.5cm}{\centering{Nent}}} & [landlord] Landlord hereby \textbf{waives the right to} any revenue that may be generated as a result of the use of the roof by Tenant or any other third - parties pursuant to the terms of the Lease during the Term.\\
& [lessee] The Lessee \textbf{will not be entitled to} a reimbursement of any part of the Rent, even if in practice the Building Capacity for which it has paid has not been utilized. \\
\hline
\multirow{2}{*}{\parbox{0.5cm}{\centering{None}}} & [lessor] For the avoidance of doubt, it is hereby clarified that wherever the word Lessor is written this means: "the Lessor and/or anyone acting on its behalf".\\
& [landlord] Other than the Purchase Agreement, this Lease represents the entire agreement and understanding between Landlord and Tenant with respect to the subject matter herein, and there are no representations, understandings, stipulations, agreements or promises not incorporated in writing herein.\\
\bottomrule
\end{tabular}
}
\caption{Sample annotated sentences for each deontic type with respect to an [Agent] and trigger annotations in \textbf{bold-face}.}\label{tab:examples}
\end{table*}
\subsection{Case Study: Red flag Detection} \label{sec:app-redflag}
Evaluation scores for the red flag detection case study are presented in Table~\ref{tab:case}.
\subsection{Annotated Examples for Deontic Types} \label{sec:example}
Samples annotations are provided in Table~\ref{tab:examples}.
\end{document}